\documentclass[10pt,twocolumn,letterpaper]{article}

\usepackage{iccv}
\usepackage{times}
\usepackage{graphicx}
\usepackage{amsmath}
\usepackage{amssymb}
\usepackage[american]{babel}
\usepackage{microtype}
\usepackage{algorithm}
\usepackage{algorithmic}
\usepackage{multirow}
\usepackage{threeparttable}

\graphicspath{{figures/}}
\DeclareMathOperator*{\argmin}{arg\,min}

\usepackage[pagebackref=true,breaklinks=true,colorlinks,bookmarks=false]{hyperref}

\iccvfinalcopy 

\def\iccvPaperID{2276} 

\begin{document}

\title{ThiNet: A Filter Level Pruning Method for Deep Neural Network Compression}

\author{Jian-Hao Luo$^1$, Jianxin Wu$^1$, and Weiyao Lin$^2$\\
	$^1$National Key Laboratory for Novel Software Technology, Nanjing University, Nanjing, China\\
	$^2$Shanghai Jiao Tong University, Shanghai, China\\
	{\tt\small luojh@lamda.nju.edu.cn, wujx2001@nju.edu.cn, wylin@sjtu.edu.cn}
}

\maketitle

\begin{abstract}
We propose an efficient and unified framework, namely ThiNet, to simultaneously accelerate and compress CNN models in both training and inference stages. We focus on the filter level pruning, i.e., the whole filter would be discarded if it is less important. Our method does not change the original network structure, thus it can be perfectly supported by any off-the-shelf deep learning libraries. We formally establish filter pruning as an optimization problem, and reveal that we need to prune filters based on statistics information computed from its next layer, not the current layer, which differentiates ThiNet from existing methods. Experimental results demonstrate the effectiveness of this strategy, which has advanced the state-of-the-art. We also show the performance of ThiNet on ILSVRC-12 benchmark. ThiNet achieves 3.31$\times$ FLOPs reduction and 16.63$\times$ compression on VGG-16, with only 0.52$\%$ top-5 accuracy drop. Similar experiments with ResNet-50 reveal that even for a compact network, ThiNet can also reduce more than half of the parameters and FLOPs, at the cost of roughly 1$\%$ top-5 accuracy drop. Moreover, the original VGG-16 model can be further pruned into a very small model with only 5.05MB model size, preserving AlexNet level accuracy but showing much stronger generalization ability.
\end{abstract}

\section{Introduction}

In the past few years, we have witnessed a rapid development of deep neural networks in the field of computer vision, from basic image classification tasks such as the ImageNet recognition challenge~\cite{Krizhevsky12NIPS,Simonyan14ICLR,He16CVPR}, to some more advanced applications, \eg, object detection~\cite{fast_R_CNN}, semantic segmentation~\cite{Noh15ICCV}, image captioning~\cite{Jia15ICCV} and many others. Deep neural networks have achieved state-of-the-art performance in these fields compared with traditional methods based on manually designed visual features. 

In spite of its great success, a typical deep model is hard to be deployed on resource constrained devices, \eg, mobile phones or embedded gadgets. A resource constrained scenario means a computing task must be accomplished with limited resource supply, such as computing time, storage space, battery power, etc. One of the main issues of deep neural networks is its huge computational cost and storage overhead, which constitute a serious challenge for a mobile device. For instance, the VGG-16 model~\cite{Simonyan14ICLR} has 138.34 million parameters, taking up more than 500MB storage space,\footnote{1 MB$\,=2^{20}\approx$ 1.048 million bytes, and 1 million is $10^6$.} and needs 30.94 billion float point operations (FLOPs) to classify a single image. Such a cumbersome model can easily exceed the computing limit of small devices. Thus, network compression has drawn a significant amount of interest from both academia and industry.

Pruning is one of the most popular methods to reduce network complexity, which has been widely studied in the model compression community. In the 1990s, LeCun~\etal~\cite{LeCun1990NIPS_brain_damage} had observed that several unimportant weights can be removed from a trained network with negligible loss in accuracy. A similar strategy was also explored in~\cite{Chechik1998}. This process resembles the biological phenomena in mammalian brain, where the number of neuron synapses has reached the peak in early childhood, followed by gradual pruning during its development. However, these methods are mainly based on the second derivative, thus are not applicable for today's deep model due to expensive memory and computation costs.

Recently, Han~\etal~\cite{Han15NIPS} introduced a simple pruning strategy: all connections with weights below a threshold are removed, followed by fine-tuning to recover its accuracy. This iterative procedure is performed several times, generating a very sparse model. However, such a non-structured sparse model can not be supported by off-the-shelf libraries, thus specialized hardwares and softwares are needed for efficient inference, which is difficult and expensive in real-world applications. On the other hand, the non-structured random connectivity ignores cache and memory access issues. As indicated in~\cite{Wen16NIPS}, due to the poor cache locality and jumping memory access caused by random connectivity, the \emph{practical acceleration} is very limited (sometimes even slows down), even though the actual sparsity is relatively high.

To avoid the limitations of non-structured pruning mentioned above, we suggest that the \emph{filter level} pruning would be a better choice. The benefits of removing the whole unimportant filter have a great deal: 1) The pruned model has no difference in network structure, thus it can be perfectly supported by any off-the-shelf deep learning libraries. 2) Memory footprint would be reduced dramatically. Such memory reduction comes not only from model parameter itself, but also from the intermediate activation, which is rarely considered in previous studies. 3) Since the pruned network structure has not be damaged, it can be further compressed and accelerated by other compression methods, \eg, the parameter quantization approach~\cite{Wu16CVPR}. 4) More vision tasks, such as object detection or semantic segmentation, can be accelerated greatly using the pruned model.

In this paper, we propose a unified framework, namely ThiNet (stands for \emph{``Thin Net''}), to prune the unimportant filters to simultaneously accelerate and compress CNN models in both training and test stages with minor performance degradation. With our pruned network, some important transfer tasks such as object detection or fine-grained recognition can run much faster (both training and inference), especially in small devices. Our main insight is that we establish \emph{a well-defined optimization problem}, which shows that \emph{whether a filter can be pruned depends on the outputs of its \emph{next} layer, not its own layer}. This novel finding differentiates ThiNet from existing methods which prune filters using statistics calculated from their own layer.

We then compare the proposed method with other state-of-the-art criteria. Experimental results show that our approach is significantly better than existing methods, especially when the compression rate is relatively high. We evaluate ThiNet on the large-scale ImageNet classification task. ThiNet achieves $3.31\times$ FLOPs reduction and $16.63\times$ compression on VGG-16 model~\cite{Simonyan14ICLR}, with only $0.52\%$ top-5 accuracy drop. The ResNet-50 model~\cite{He16CVPR} has less redundancy compared with classic CNN models. ThiNet can still reduce $2.26\times$ FLOPs and $2.06\times$ parameters with roughly $1\%$ top-5 accuracy drop. To explore the limits of ThiNet, we show that the original VGG-16 model can even be pruned into 5.05MB, but still preserving AlexNet level accuracy.

In addition, we also explore the performance of ThiNet in a more practical task, \ie, transfer learning on small-scale datasets. Experimental results demonstrate the excellent effectiveness of ThiNet, which achieves the best trade-off between model size and accuracy.

The key advantages and major contributions of this paper can be summarized as follows.
\begin{itemize} \setlength{\itemsep}{-2pt}
	\item We propose a simple yet effective framework, namely ThiNet, to simultaneously accelerate and compress CNN models. ThiNet shows significant improvements over existing methods on numerous tasks.
	\item We formally establish filter pruning as an optimization problem, and reveal that we need to prune filters using statistics information computed from its next layer, not the current layer, which differentiates ThiNet from existing methods.
	\item In experiments, the VGG-16 model can be pruned into 5.05MB, showing promising generalization ability on transfer learning. Higher accuracy could be preserved with a more accurate model using ThiNet.
\end{itemize}

\section{Related work}

Many researchers have found that deep models suffer from heavy over-parameterization. For example, Denil \etal~\cite{Denil13NIPS} demonstrated that a network can be efficiently reconstructed with only a small subset of its original parameters. However, this redundancy seems \emph{necessary} during model training, since the highly non-convex optimization is hard to be solved with current techniques~\cite{Denton14NIPS, Hinton12arxiv}. Hence, there is a great need to reduce model size \emph{after} its training.

Some methods have been proposed to pursuit a balance between model size and accuracy. Han \etal~\cite{Han15NIPS} proposed an iterative pruning method to remove the redundancy in deep models. Their main insight is that small-weight connectivity below a threshold should be discarded. In practice, this can be aided by applying $\ell_1$ or $\ell_2$ regularization to push connectivity values becoming smaller. The major weakness of this strategy is the loss of universality and flexibility, thus seems to be less practical in the real applications.

In order to avoid these weaknesses, some attention has been focused on the group-wise sparsity. Lebedev and Lempitsky~\cite{Lebedev16CVPR} explored group-sparse convolution by introducing the group-sparsity regularization to the loss function, then some entire groups of weights would shrink to zeros, thus can be removed. Similarly, Wen~\etal~\cite{Wen16NIPS} proposed the Structured Sparsity Learning (SSL) method to regularize filter, channel, filter shape and depth structures. In spite of their success, the original network structure has been destroyed. As a result, some dedicated libraries are needed for an efficient inference speed-up.

In line with our work, some filter level pruning strategies have been explored too. The core is to evaluate neuron importance, which has been widely studied in the community~\cite{CAM_Zhou_NIPS, Grad_CAM, Li17ICLR, Hu16arxiv, Molchanov17ICLR}. A simplest possible method is based on the magnitude of weights. Li~\etal~\cite{Li17ICLR} measured the importance of each filter by calculating its absolute weight sum. Another practical criterion is to measure the sparsity of activations after the ReLU function. Hu~\etal~\cite{Hu16arxiv} believed that if most outputs of some neurons are zero, these activations should be expected to be redundant. They compute the Average Percentage of Zeros (APoZ) of each filter as its importance score. These two criteria are simple and straightforward, but not directly related to the final loss. Inspired by this observation, Molchanov~\etal~\cite{Molchanov17ICLR} adopted Taylor expansion to approximate the influence to loss function induced by removing each filter.

Beyond pruning, there are also other strategies to obtain small CNN models. One popular approaches is parameter quantization~\cite{Gong14arxiv, Chen15ICML, Wu16CVPR, Han16ICLR}. Low-rank approximation is also widely studied~\cite{Denton14NIPS, Sindhwani15NIPS}. Note that these methods are complementary to filter pruning, which can be combined with ThiNet for further improvement.

\section{ThiNet}

In this section, we will give a comprehensive introduction to our filter level pruning approach: ThiNet. First, the overall framework will be presented. Next, a more detailed description of our selection algorithm would be presented. Finally, we will show our pruning strategy, which takes both efficiency and effectiveness into consideration. 

\subsection{Framework of ThiNet}

Pruning is a classic method used for reducing model complexity. Although vast differences exist (such as different criteria in selecting what should be pruned), the overall framework is similar in pruning filters inside a deep neural network. It can be summarized in one sentence: evaluate the importance of each neuron, remove those unimportant ones, and fine-tune the whole network.

\begin{figure}
	\centering
	\includegraphics[width=1.0\linewidth]{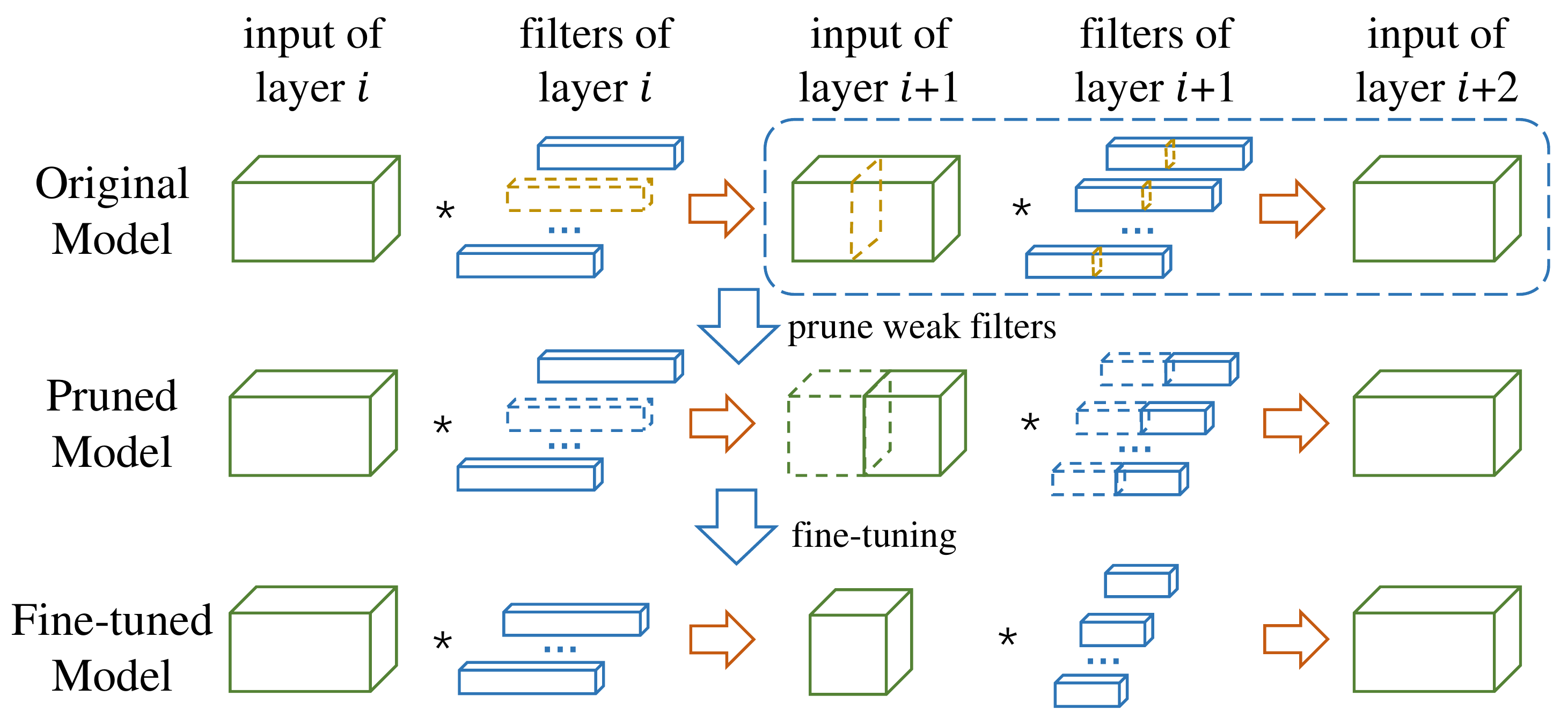}
	\caption{Illustration of ThiNet. First, we focus on the dotted box part to determine several weak channels and their corresponding filters (highlighted in yellow in the first row). These channels (and their associated filters) have little contribution to the overall performance, thus can be discarded, leading to a pruned model. Finally, the network is fine-tuned to recover its accuracy. (This figure is best viewed in color.)}
	\label{framework}
\end{figure}

This framework is illustrated in Figure~\ref{framework}. In the next sub-section, we will focus on the dotted box part to introduce our data-driven channel selection method, which determines the channels (and their associated filters) that are to be pruned away.

Given a pre-trained model, it would be pruned layer by layer with a predefined compression rate. We summarize our framework as follows:
\begin{enumerate} \setlength{\itemsep}{-2pt}
	\item \textbf{Filter selection.} Unlike existing methods that use layer $i$'s statistics to guide the pruning of layer $i$'s filters, we use layer $i+1$ to guide the pruning in layer $i$. The key idea is: \emph{if we can use a subset of channels in layer $(i+1)$'s input to approximate the output in layer $i+1$, the other channels can be safely removed} from the input of layer $i+1$. Note that one channel in layer $(i+1)$'s input is produced by one filter in layer $i$, hence we can safely prune the corresponding filter in layer $i$.
	
	\item \textbf{Pruning.} Weak channels in layer $(i+1)$'s input and their corresponding filters in layer $i$ would be pruned away, leading to a much smaller model. Note that, the pruned network has exactly the same structure but with fewer filters and channels. In other words, the original wide network is becoming much thinner. That is why we call our method ``ThiNet''.
	
	\item \textbf{Fine-tuning.} Fine-tuning is a necessary step to recover the generalization ability damaged by filter pruning. But it will take very long for large datasets and complex models. For time-saving considerations, we fine-tune one or two epochs after the pruning of one layer. In order to get an accurate model, more additional epochs would be carried out when all layers have been pruned.

	\item \textbf{Iterate to step 1 to prune the next layer.}
\end{enumerate}
	
\subsection{Data-driven channel selection}

We use a triplet $\langle \mathcal{I}_i, \mathcal{W}_i, \ast\rangle$ to denote the convolution process in layer $i$, where $\mathcal{I}_i \in \mathbb{R}^{C\times H \times W}$ is the input tensor, which has $C$ channels, $H$ rows and $W$ columns. And $\mathcal{W}_i \in \mathbb{R}^{D\times C \times K \times K}$ is a set of filters with $K \times K$ kernel size, which generates a new tensor with $D$ channels.

Our goal is to remove some unimportant filters in $\mathcal{W}_i$. Note that, if a filter in $\mathcal{W}_i$ is removed, its corresponding channel in $\mathcal{I}_{i+1}$ and $\mathcal{W}_{i+1}$ would also be discarded. However, since the filter number in layer $i+1$ has not been changed, the size of its output tensor, \ie, $\mathcal{I}_{i+2}$, would be kept exactly the same. Inspired by this observation, we believe that if we can remove several filters that has little influence on $\mathcal{I}_{i+2}$ (which is also the output of layer $i+1$), it would have little influence on the overall performance too. In other words, minimizing the reconstruction error of $\mathcal{I}_{i+2}$ is closely related to the network's classification performance.

\subsubsection{Collecting training examples} \label{subsubsec_training_examples}

\begin{figure}[t]
  \centering
  \includegraphics[width=0.9\linewidth]{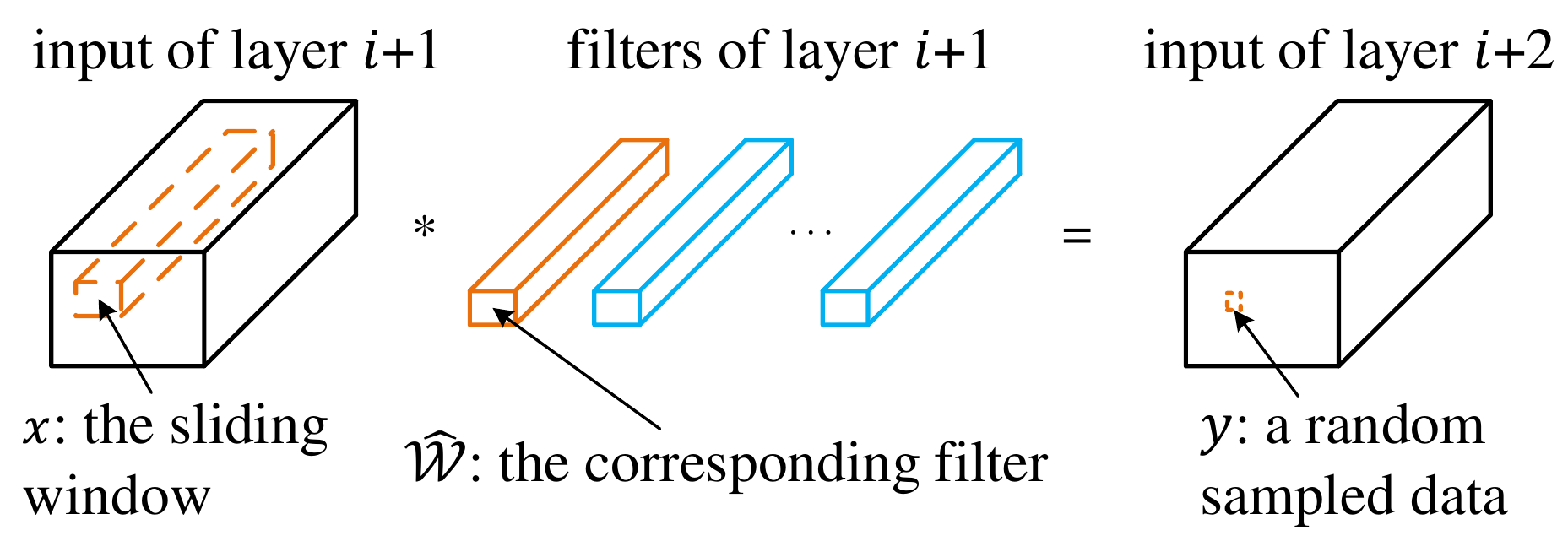}
  \caption{Illustration of data sampling and variables' relationship.}
  \label{Data_Sampling}
\end{figure}

In order to determine which channel can be removed safely, a training set used for importance evaluation would be collected. As illustrated in Figure~\ref{Data_Sampling}, an element, denoted by $y$, is randomly sampled from the tensor $\mathcal{I}_{i+2}$ (before ReLU). A corresponding filter $\widehat{\mathcal{W}} \in \mathbb{R}^{C \times K \times K}$ and sliding window $x\in \mathbb{R}^{C\times K \times K}$ (after ReLU) can also be determined according to its location. Here, some index notations are omitted for a clearer presentation. Normally, the convolution operation can be computed with a corresponding bias $b$ as follows:
\begin{equation}
\label{Eq. 1}
	y = \sum_{c=1}^C\sum_{k_1=1}^K\sum_{k_2=1}^K\widehat{\mathcal{W}}_{c,k_1,k_2}\times x_{c,k_1,k_2}+b.
\end{equation}
Now, if we further define:
\begin{equation}
	\hat{x}_c=\sum_{k_1=1}^K\sum_{k_2=1}^K\widehat{\mathcal{W}}_{c,k_1,k_2}\times x_{c,k_1,k_2} ,
\end{equation}
Eq.~\ref{Eq. 1} can be simplified as:
\begin{equation}
\label{Eq. 3}
	\hat{y} = \sum_{c=1}^C\hat{x}_c ,
\end{equation}
in which $\hat{y} = y - b$. It is worthwhile to keep in mind that $\hat{x}$ and $\hat{y}$ are random variables whose instantiations require fixed spatial locations indexed by $c$, $k_1$ and $k_2$. A key observation is that channels in $\hat{\mathbf{x}}=(\hat{x}_1,\hat{x}_2,\dots,\hat{x}_C)$ is independent: $\hat{x}_c$ \emph{only depends} on $x_{c,:,:}$, which has no dependency relationship with $x_{c',:,:}$, if $c' \neq c$.

In other words, if we can find a subset $S \subset \{1,2,\ldots, C\}$ and the equality
\begin{equation}
\label{Eq. 4}
	\hat{y} = \sum_{c \in S}\hat{x}_c
\end{equation}
always holds, then we do not need any $\hat{x}_c$ if $c \notin S$ and these variables can be \emph{safely removed without changing the CNN model's result}.

Of course, Eq.~\ref{Eq. 4} cannot always be true for all instances of the random variables $\hat{\mathbf{x}}$ and $\hat{y}$. However, we can manually extract instances of them to find a subset $S$ such that Eq.~\ref{Eq. 4} is approximately correct.

Given an input image, we first apply the CNN model in the forward run to find the input and output of layer $i+1$. Then for any feasible $(c,k_1,k_2)$ triplet, we can obtain a $C$-dimensional vector variable $\mathbf{\hat{x}}=\{\hat{x}_1, \hat{x}_2, \ldots, \hat{x}_C\}$ and a scalar value $\hat{y}$ using Eq.~\ref{Eq. 1} to Eq.~\ref{Eq. 3}. Since $\mathbf{\hat{x}}$ and $\hat{y}$ can be viewed as random variables, more instances can be sampled by choosing different input images, different channels, and different spatial locations.

\subsubsection{A greedy algorithm for channel selection}

Now, given a set of $m$ (the product of number of images and number of locations) training examples $\{(\mathbf{\hat{x}}_i, \hat{y}_i)\}$, the original channel selection problem becomes the following optimization problem:
\begin{equation}
\label{object1}
	\begin{aligned}
	\argmin_S  & \sum_{i=1}^{m}\left(\hat{y}_i-\sum_{j\in S}\mathbf{\hat{x}}_{i,j}\right)^2\\
	\text{s.t.}\quad & |S|=C\times r, \quad S \subset \{1,2,\ldots, C\}.
	\end{aligned}
\end{equation}
Here, $|S|$ is the number of elements in a subset $S$, and $r$ is a pre-defined compression rate (\ie, how many channels are preserved). Equivalently, let $T$ be the subset of removed channels (\ie, $S\cup T=\{1,2,\ldots,C\}$ and $S\cap T=\emptyset$), we can minimize the following alternative objective:
\begin{equation}
\label{object2}
\begin{aligned}
\argmin_T & \sum_{i=1}^{m}\left(\sum_{j\in T}\mathbf{\hat{x}}_{i,j}\right)^2\\
\text{s.t.}\quad & |T|=C\times (1-r), \quad T \subset \{1,2,\ldots, C\}.
\end{aligned}
\end{equation}
Eq.~\ref{object2} is equivalent to Eq.~\ref{object1}, but has faster speed because $|T|$ is usually smaller than $|S|$. Solving Eq.~\ref{object2} is still NP hard, thus we use a greedy strategy (illustrated in algorithm~\ref{alg:1}). We add one element to $T$ at a time, and choose the channel leading to the smallest objective value in the current iteration.

\begin{algorithm}[t]
	\caption{A greedy algorithm for minimizing Eq.~\ref{object2}}  
	\label{alg:1}  
	\begin{algorithmic} [1]
		\REQUIRE ~
		Training set $\{(\mathbf{\hat{x}}_i, \hat{y}_i)\}$, and compression rate $r$
		\ENSURE
		The subset of removed channels: $T$
		\STATE {$T\leftarrow \emptyset;$ $I\leftarrow\{1,2,\ldots,C\};$}
		\WHILE {$\left| T \right|<C\times (1-r)$}  
		  \STATE {$min\_value\leftarrow+\infty;$}\\
		  \FOR {each item $i \in I$}
		    \STATE {$tmpT\leftarrow T \cup \{i\};$}
		    \STATE {compute $value$ from Eq.~\ref{object2} using $tmpT;$}
		    \IF {$value<min\_value$}
		      \STATE{$min\_value\leftarrow value;$ $min\_i\leftarrow i;$}
		    \ENDIF
		  \ENDFOR
		\STATE{move $min\_i$ from $I$ into $T;$}
		\ENDWHILE
	\end{algorithmic}  
\end{algorithm}

Obviously, this greedy solution is sub-optimal. But the gap can be compensated by fine-tuning. We have also tried some other sophisticated algorithms, such as sparse coding (specifically, the homotopy method~\cite{Donoho2008L1}). However, our simple greedy approach has better performance and faster speed according to our experiments.

\subsubsection{Minimize the reconstruction error}

So far, we have obtained the subset $T$ such that the $n$-th channel in each filter of layer $i+1$ can be safely removed if $n \in T$. Hence, the corresponding filters in the previous layer $i$ can be pruned too. 

Now we will further minimize the reconstruction error (\cf Eq.~\ref{object1}) by weighing the channels, which can be defined as:
\begin{equation}
\label{Eq. 7}
	\mathbf{\hat{w}}=\argmin_\mathbf{w}\sum_{i=1}^{m}(\hat{y}_i- \mathbf{w}^\mathrm{T}\mathbf{\hat{x}}^*_i)^2,
\end{equation}
where $\mathbf{\hat{x}}^*_i$ indicates the training examples after channel selection. Eq.~\ref{Eq. 7} is a classic linear regression problem, which has a unique closed-form solution using the ordinary least squares approach: $\mathbf{\hat{w}} = (\mathbf{X}^\mathrm{T}\mathbf{X})^{-1}\mathbf{X}^\mathrm{T}\mathbf{y}$.

Each element in $\mathbf{\hat{w}}$ can be regarded as a scaling factor of corresponding filter channel such that $\mathcal{W}_{:,i,:,:}=\hat{w}_i \mathcal{W}_{:,i,:,:}$. From another point of view, this scaling operation provides a better initialization for fine-tuning, hence the network is more likely to reach higher accuracy. 

\subsection{Pruning strategy} \label{subsec_pruning_strategy}

There are mainly two types of different network architectures: the traditional convolutional/fully-connected architecture, and recent structural variants. The former is represented by AlexNet~\cite{Krizhevsky12NIPS} or VGGNet~\cite{Simonyan14ICLR}, while the latter mainly includes some recent networks like GoogLeNet~\cite{Szegedy15CVPR} and ResNet~\cite{He16CVPR}. The main difference between these two types is that more recent networks usually replace the FC (fully-connected) layers with a global average pooling layer~\cite{NIN, CAM_Zhou_NIPS}, and adopt some novel network structures like Inception in GoogLeNet or residual blocks in ResNet.

We use different strategies to prune these two types of networks. For VGG-16,  we notice that more than 90\% FLOPs exist in the first 10 layers (conv1-1 to conv4-3), while the FC layers contribute nearly 86.41\% parameters. Hence, we prune the first 10 layers for acceleration consideration, but replace the FC layers with a global average pooling layer. Although the proposed method is also valid for FC layers, we believe removing them is simpler and more efficient. 

\begin{figure}[t]
	\centering
	\includegraphics[width=0.95\linewidth]{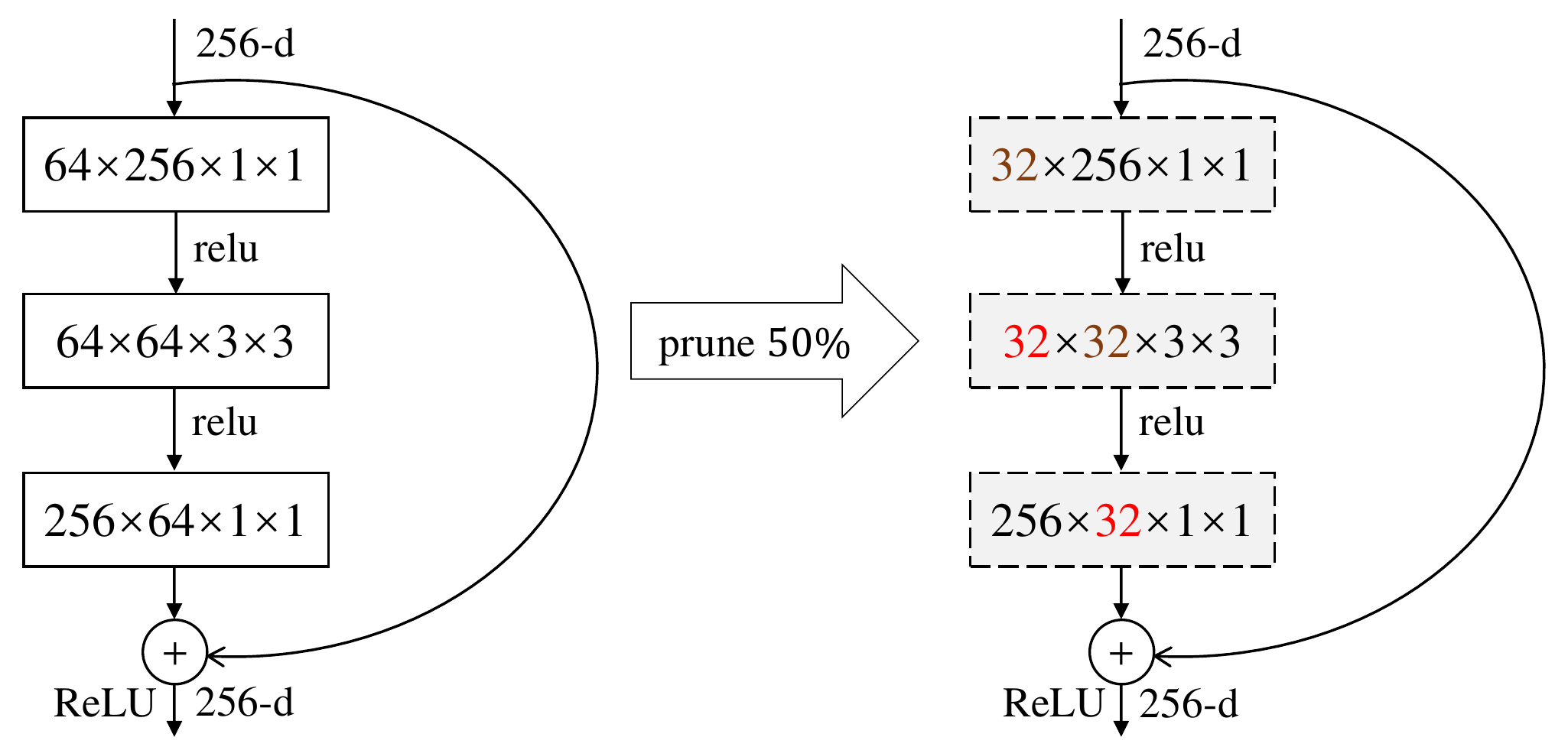}
	\caption{Illustration of the ResNet pruning strategy. For each residual block, we only prune the first two convolutional layers, keeping the block output dimension unchanged.}
	\label{ResNet}
\end{figure}

For ResNet, there exist some restrictions due to its special structure. For example, the channel number of each block in the same group needs to be consistent in order to finish the sum operation (see~\cite{He16CVPR} for more details). Thus it is hard to prune the last convolutional layer of each residual block directly. Since most parameters are located in the first two layers, pruning the first two layers is a good choice, which is illustrated in Figure~\ref{ResNet}.

\section{Experiments}

We empirically study the performance of ThiNet in this section. First, a comparison among several different filter selection criteria would be presented. Experimental results show that our method is significantly better than others. Then, we would report the performance on ILSCVR-12~\cite{ILSVRC15}. Two widely used networks are pruned: VGG-16~\cite{Simonyan14ICLR} and ResNet-50~\cite{He16CVPR}. Finally, we focus on a more practical scenario to show the advantages of ThiNet. All the experiments are conducted within Caffe~\cite{jia2014caffe}. 

\subsection{Different filter selection criteria} \label{exp_1}

There exist some heuristic criteria to evaluate the importance of each filter in the literature. We compare our selection method with two recently proposed criteria to demonstrate the effectiveness of our evaluation criterion. These criteria are briefly summarized as follows:
\begin{itemize} \setlength{\itemsep}{-2pt}
	\item \textbf{Weight sum}~\cite{Li17ICLR}. Filters with smaller kernel weights tend to produce weaker activations. Thus, in this strategy the absolute sum of each filter is calculated as its importance score: $s_i = \sum|\mathcal{W}(i,:,:,:)|$.
	
	\item \textbf{APoZ (Average Percentage of Zeros)}~\cite{Hu16arxiv}. This criterion calculates the sparsity of each channel in output activations as its importance score: $s_i=\frac{1}{|\mathcal{I}(i,:,:)|}\sum\sum\mathbb{I}(\mathcal{I}(i,:,:)==0)$, where $|\mathcal{I}(i,:,:)|$ is the elements number in $i$-th channel of tensor $\mathcal{I}$ (after ReLU), and $\mathbb{I}(\cdot)$ denotes the indicator function.
\end{itemize}

To compare these different selection methods, we evaluate their performance on the widely used fine-grained dataset: CUB-200~\cite{CUB_200_2011}, which contains 11,788 images of 200 different bird species (5994/5794 images for training/test, respectively). Except for labels, no additional supervised information (\eg, bounding box) is used. 

Following the pruning strategy in Section~\ref{subsec_pruning_strategy}, all the FC layers in VGG-16 are removed, and replaced with a global average pooling layer, and fine-tuned on new datasets. Starting from this fine-tuned model, we then prune the network layer by layer with different compression rate. Each pruning is followed by one epoch fine-tuning, and 12 epochs are performed in the final layer to improve accuracy. This procedure is repeated several times with different channel selection strategies. Due to the random nature of ThiNet, we repeated our method 4 times and report the averaged result. For a fair comparison, all the settings are kept the same, except the selection method.

\begin{figure} 
  \centering 
  \includegraphics[width=0.95\columnwidth]{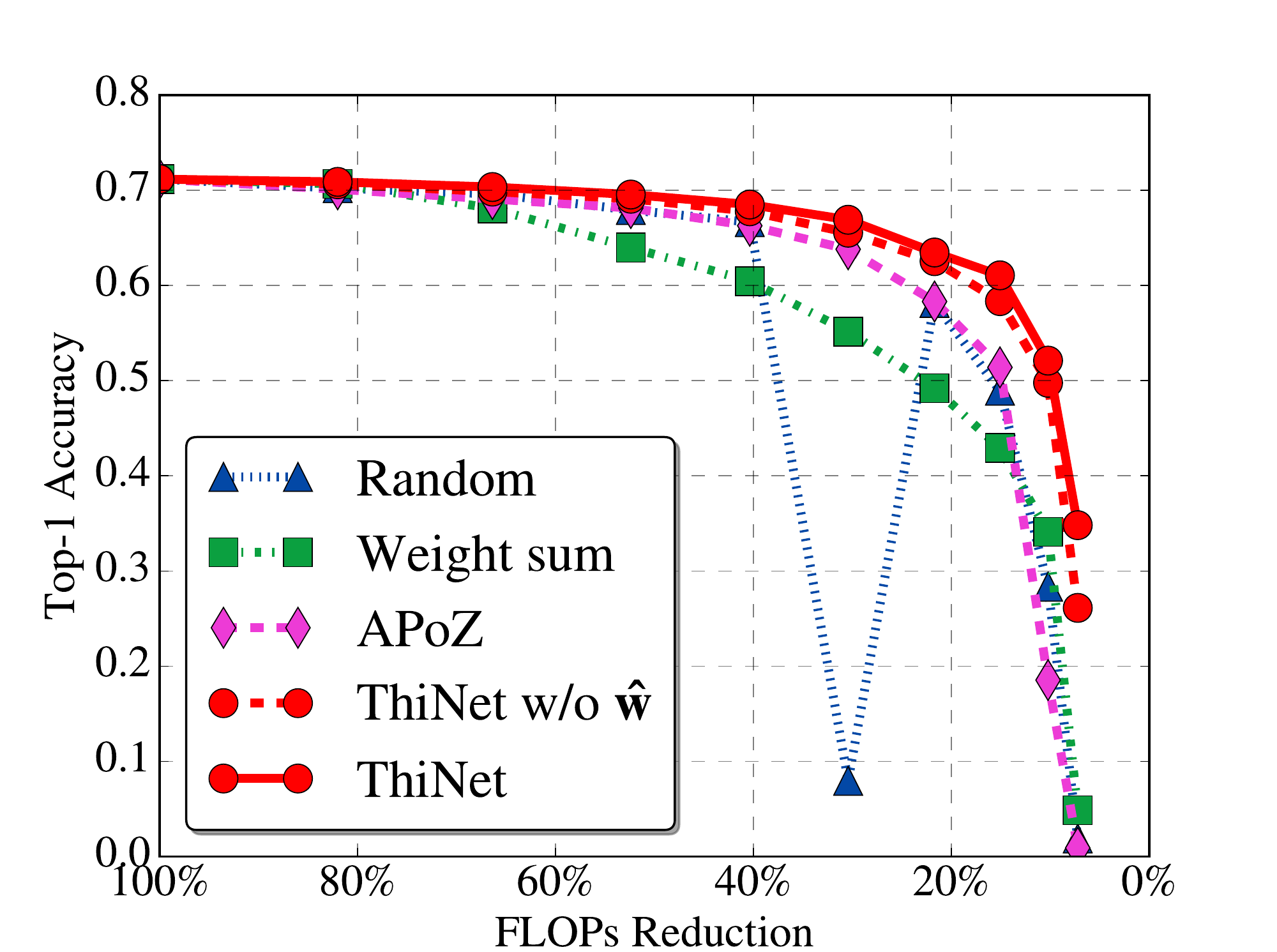}
  \caption{Performance comparison of different channel selection methods: the VGG-16-GAP model pruned on CUB-200 with different compression rates. (This figure is best viewed in color and zoomed in.)}
  \label{fig} 
\end{figure}

Figure~\ref{fig} shows the pruning results on the CUB bird dataset. We also evaluated the performance of random selection with the same pruning strategy. In addition, another version of ThiNet without least squares (denoted by ``ThiNet w/o $\mathbf{\hat{w}}$'') is also evaluated to demonstrate the effectiveness of least squares in our method. Obviously, ThiNet achieves consistently and significantly higher accuracy compared with other selection methods. 

One interesting result is: random selection shows pretty good performance, even better than heuristic criteria in some cases. In fact, according to the property of distributed representations (\ie, each concept is represented by many neurons; and, each neuron participates in the representation of many concepts~\cite{Hinton86CogSci, Bengiobeng13TPAMI}), randomly selected channels may be quite powerful in theory. However, this criterion is not robust. As shown in Figure~\ref{fig}, it can lead to very bad result and the accuracy is very low after all layers are compressed. Thus, random selection is not applicable in practice.

Weight sum has pretty poor accuracy on CUB-200. This result is reasonable, since it only takes the magnitude of kernel weights into consideration, which is not directly related to the final classification accuracy. In fact, small weights could still have large impact on the loss function. When we discard a large number of small filters at the same time, the final accuracy can be damaged greatly. For example, if we removed $60\%$ filters in conv1-1 using the small weight criterion, the top-1 accuracy is only $40.99\%$ (before fine-tuning), while random criterion is $51.26\%$. By contrast, our method (ThiNet w/o w) can reach $68.24\%$, and even $70.75\%$ with least squares (ThiNet). The accuracy loss of weight sum is so large that fine-tuning cannot completely recover it from the drop. 

In contrast, our method shows much higher and robust results. The least squares approach does indeed aid to get a better weight initialization for fine-tuning, especially when the compression rate is relatively high.

\subsection{VGG-16 on ImageNet} \label{sec_vgg16_ImageNet}

We now evaluate the performance of the proposed ThiNet method on large-scale ImageNet classification task. The ILSCVR-12 dataset~\cite{ILSVRC15} consists of over one million training images drawn from 1000 categories. We randomly select 10 images from each category in the training set to comprise our evaluation set (\ie, collected training examples for channel selection). And for each input image, 10 instances are randomly sampled with different channels and different spatial locations as described in section~\ref{subsubsec_training_examples}. Hence, there are in total 100,000 training samples used for finding the optimal channel subset via Algorithm~\ref{alg:1}. We compared several different choices of image and location number, and found that the current choice (10 images per class and 10 locations per image) is enough for neuron importance evaluation. Finally, top-1 and top-5 classification performance are reported on the 50k standard validation set, using the single-view testing approach (central patch only).

During fine-tuning, images are resized to $256\times256$, then $224\times 224$ random cropping is adopted to feed the data into network. Horizontal flip is also used for data augmentation. At the inference stage, we center crop the resized images to $224\times 224$. No more tricks are used here. The whole network is pruned layer by layer and fine-tuned in one epoch with $10^{-3}$ learning rate. Since the last layer of each group (\ie, conv1-2, conv2-2, conv3-3) is more important (pruning these layers would lead to a big accuracy drop), we fine-tune these layers with additional one epoch of $10^{-4}$ learning rate to prevent accuracy drop too much. When pruning the last layer, more epochs (12 epochs) are adopted to get an accurate result with learning rate varying from $10^{-3}$ to $10^{-5}$. We use SGD with mini-batch size of 128, and other parameters are kept the same as the original VGG paper~\cite{Simonyan14ICLR}.

We summarize the performance of the ThiNet approach in Table~\ref{vgg16}. Here, ``ThiNet-Conv'' refers to the model in which only the first 10 convolutional layers are pruned with compression rate 0.5 (\ie, half of the filters are removed in each layer till conv4-3) as stated above. Because some useless filters are discarded, the pruned model can even outperform the original VGG-16 model. However, if we train this model from scratch, the top-1/top-5 accuracy are only 67.00$\%$/87.45$\%$ respectively, which is much worse than our pruned network. Then the FC layers are removed, replaced with a GAP (global average pooling) layer and fine-tuned in 12 epochs with the same hyper-parameters, which is denoted by ``ThiNet-GAP''. The classification accuracy of GAP model is slightly lower than the original model, since the model size has been reduced dramatically. Further reduction can be obtained with a higher compression rate (denoted by ``ThiNet-Tiny''), which would be discussed later.

The actual speed-up of ThiNet is also reported. We test the forward/backward running time of each model using the official ``time'' command in Caffe. This evaluation is conducted on one M40 GPU with batch size 32 accelerated by cuDNN v5.1. Since convolution operations dominate the computational costs of VGG-16, reducing FLOPs would accelerate inference speed greatly, which is shown in Table~\ref{vgg16}.

\begin{table}
	\caption{Pruning results of VGG-16 on ImageNet using ThiNet. Here, M/B means million/billion ($10^6$/$10^9$), respectively; f./b. denotes the forward/backward timing in milliseconds tested on one M40 GPU with batch size 32.}
	\label{vgg16}
	\footnotesize
	\setlength{\tabcolsep}{2.2pt} 
	\centering
	\begin{threeparttable}
	\begin{tabular}{|l|c|c|c|c|c|}
		\hline
		Model & Top-1 & Top-5 & \#Param. & \#FLOPs\tnote{1} & f./b. (ms)\\
		\hline \hline
		Original\tnote{2} & 68.34\% & 88.44\% & 138.34M & 30.94B & 189.92/407.56 \\
		ThiNet-Conv & 69.80\% & 89.53\% & 131.44M & 9.58B & 76.71/152.05 \\
		Train from scratch  & 67.00\% & 87.45\% & 131.44M & 9.58B & 76.71/152.05 \\
		ThiNet-GAP & 67.34\% & 87.92\% & 8.32M & 9.34B & 71.73/145.51 \\
		\hline \hline
		ThiNet-Tiny & 59.34\% & 81.97\% & 1.32M & 2.01B & 29.51/55.83 \\
		SqueezeNet\cite{Iandola16Squeeze} & 57.67\% & 80.39\% & 1.24M & 1.72B & 37.30/68.62 \\    
		\hline
	\end{tabular}
	\begin{tablenotes}
		\footnotesize
		\item[1] In this paper, we only consider the FLOPs of convolution operations, which is commonly used for computation complexity comparison.
		\item[2] For a fair comparison, the accuracy of original VGG-16 model is evaluated on resized center-cropped images using pre-trained model as adopted in~\cite{Han15NIPS, Hu16arxiv}. The same strategy is also used in ResNet-50.
	\end{tablenotes}
	\end{threeparttable}
\end{table}

We then compare our approach with several state-of-the-art pruning methods on the VGG-16 model, which is shown in Table~\ref{vgg16_compare}. These methods also focus on filter-level pruning, but with totally different selection criteria.

\begin{table}
	\caption{Comparison among several state-of-the-art pruning methods on the VGG-16 network. Some exact values are not reported in the original paper and cannot be computed, thus we use $\approx$ to denote the approximation value.}
	\label{vgg16_compare}
	\footnotesize
	\setlength{\tabcolsep}{4.5pt} 
	\centering
	\begin{tabular}{|c|c|c|c|c|}
		\hline
		Method &Top-1 Acc.&Top-5 Acc.&\#Param. $\downarrow$ &\#FLOPs $\downarrow$\\
		\hline \hline
		APoZ-1~\cite{Hu16arxiv} & -2.16\% & -0.84\% & $2.04\times$ & $\approx1\times$ \\
		APoZ-2~\cite{Hu16arxiv} & +1.81\% & +1.25\%  & $2.70\times$ & $\approx1\times$\\ 
		Taylor-1~\cite{Molchanov17ICLR} & -- & -1.44\% & $\approx1\times$ & $2.68\times$ \\
		Taylor-2~\cite{Molchanov17ICLR} & -- & -3.94\% & $\approx1\times$ & $3.86\times$ \\
		ThiNet-WS~\cite{Li17ICLR} & +1.01\% & +0.69\% &  $1.05\times$ & $3.23\times$ \\
		\hline \hline
		ThiNet-Conv & +1.46\% & +1.09\% & $1.05\times$ & $3.23\times$ \\
		ThiNet-GAP & -1.00\% & -0.52\% & $16.63\times$ & $3.31\times$ \\
		\hline
	\end{tabular}
\end{table}

APoZ~\cite{Hu16arxiv} aims to reduce parameter numbers, but its performance is limited. APoZ-1 prunes few layers (conv4, conv5 and the FC layers), but leads to significant accuracy degradation. APoZ-2 then only prunes conv5-3 and the FC layers. Its accuracy is improved but this model almost does not reduce the FLOPs. Hence, there is a great need for compressing convolution layers. 

In contrast, Molchanov \etal ~\cite{Molchanov17ICLR} pay their attention to model acceleration, and only prune the convolutional layers. They think a filter can be removed safely if it has little influence on the loss function. But the calculating procedure can be very time-consuming, thus they use Taylor expansion to approximate the loss change. Their motivation and goals are similar to ours, but with totally different selection criterion and training framework. As shown in Table~\ref{vgg16_compare}, the ThiNet-Conv model is significantly better than Taylor method. Our model can even improve classification accuracy with more FLOPs reduction.

As for weight sum~\cite{Li17ICLR}, they have not explored its performance on VGG-16. Hence we simply replace our selection method with weight sum in the ThiNet framework, and report the final accuracy denoted by ``ThiNet-WS''. All the parameters are kept the same except for selection criterion. Note that different fine-tuning framework may lead to very different results. Hence, the accuracy may be different if Li \etal~\cite{Li17ICLR} had done this using their own framework. Because the rest setups are the same, it is fair to compare ThiNet-WS and ThiNet, and ThiNet has obtained better results.

To explore the limits of ThiNet, we prune VGG-16 with a larger compression rate 0.25, achieving $16\times$ parameters reduction in convolutional layers. The conv5 layers are also pruned to get a smaller model. As for conv5-3, which is directly related to the final feature representation, we only prune half of the filters for accuracy consideration. 

Using these smaller compression ratios, we train a very small model. Denoted as ``ThiNet-Tiny'' in Table~\ref{vgg16}, it only takes 5.05MB disk space (1MB=$2^{20}$ bytes) but still has AlexNet-level accuracy (the top-1/top-5 accuracy of AlexNet is 57.2\%/80.3\%, respectively). ThiNet-Tiny has exactly the same level of model complexity as the recently proposed compact network: SqueezeNet~\cite{Iandola16Squeeze}, but showing high accuracy. Although ThiNet-Tiny needs more FLOPs, its \emph{actual speed is even faster than SqueezeNet} because it has a much simpler network structure. SqueezeNet adopts a special structure, namely the Fire module, which is parameter efficient but relies on manual network structure design. In contrast, ThiNet is a unified framework, and higher accuracy would be obtained if we start from a more accurate model. 

\subsection{ResNet-50 on ImageNet}
We also explore the performance of ThiNet on the recently proposed powerful CNN architecture: ResNet~\cite{He16CVPR}. We select ResNet-50 as the representative of the ResNet family, which has exactly the same architecture and little difference with others.

Similar to VGG-16, we prune ResNet-50 from block 2a to 5c iteratively. Except for filters, the corresponding channels in batch-normalization layer are also discarded. After pruning, the model is fine-tuned in one epoch with fixed learning rate $10^{-4}$. And 9 epochs fine-tuning with learning rate changing from $10^{-3}$ to $10^{-5}$ is performed at the last round to gain a higher accuracy. Other parameters are kept the same as our VGG-16 pruning experiment.

\begin{table}
	\caption{Overall performance of pruning ResNet-50 on ImageNet via ThiNet with different compression rate. Here, M/B means million/billion respectively, f./b. denotes the forward/backward speed tested on one M40 GPU with batch size 32.}
	\label{resnet50}
	\footnotesize
	\setlength{\tabcolsep}{4.5pt}
	\centering
	\begin{tabular}{|c|c|c|c|c|c|}
		\hline
		Model & Top-1 & Top-5 & \#Param. & \#FLOPs & f./b. (ms)\\
		\hline \hline
		Original & 72.88\% & 91.14\% & 25.56M & 7.72B & 188.27/269.32 \\
		ThiNet-70 & 72.04\% & 90.67\% & 16.94M & 4.88B & 169.38/243.37 \\
		ThiNet-50 & 71.01\% & 90.02\% & 12.38M & 3.41B & 153.60/212.29 \\
		ThiNet-30 & 68.42\% & 88.30\% & 8.66M & 2.20B & 144.45/200.67 \\
		\hline
	\end{tabular}
\end{table}

Because ResNet is a recently proposed model, the literature lack enough works that compress this network. We report the performance of ThiNet on pruning ResNet-50, which is shown in Table~\ref{resnet50}. We prune this model with 3 different compression rates (preserve 70\%, 50\%, 30\% filters in each block respectively). Unlike VGG-16, ResNet is more compact. There exists less redundancy, thus pruning a large amount of filters seems to be more challenging. In spite of this, our method ThiNet-50 can still prune more than half of the parameters with roughly 1\% top-5 accuracy drop. Further pruning can also be carried out, leading to a much smaller model at the cost of more accuracy loss.

However, reduced FLOPs can not bring the same level of acceleration in ResNet. Due to the structure constraints of ResNet-50, non-tensor layers (\eg, batch normalization and pooling layers) take up more than 40\% of the inference time on GPU. Hence, there is a great need to accelerate these non-tensor layers, which should be explored in the future.

In this experiment, we only prune the first two layers of each block in ResNet for simplicity, leaving the block output and projection shortcuts unchanged. Pruning these parts would lead to further compression, but can be quite difficult, if not entirely impossible. And this exploration seems to be a promising extension for the future work.

\subsection{Domain adaptation ability of the pruned model} \label{subsec_domain}
One of the main advantages of ThiNet is that we have not changed network structure, thus a model pruned on ImageNet can be easily transfered into other domains. 

To help us better understand this benefit, let us consider a more practical scenario: get a small model on a domain-specific dataset. This is a very common requirement in the real-world applications, since we will not directly apply ImageNet models in a real application. To achieve this goal, there are two feasible strategies: starting from a pre-trained ImageNet model then prune on the new dataset, or train a small model from scratch. In this section, we argue that it would be a better choice if we fine-tune an already pruned model which is compressed on ImageNet. 

These strategies are compared on two different domain-specific dataset: CUB-200~\cite{CUB_200_2011} for fine-grained classification and Indoor-67~\cite{Indoor67} for scene recognition. We have introduced CUB-200 in section~\ref{exp_1}. As for Indoor-67, we follow the official train/test split (5360 training and 1340 test images) to organize this dataset. All the models are fine-tuned with the same hyper-parameters and epochs for a fair comparison. Their performance is shown in Table~\ref{domain_adaption}.
\begin{table}
	\caption{Comparison of different strategies to get a small model on CUB-200 and Indoor-67. ``FT'' stands for ``Fine Tune''.}
	\footnotesize
	\label{domain_adaption}
	\setlength{\tabcolsep}{6.5pt} 
	\centering
	\begin{tabular}{|c|c|c|c|c|}
		\hline
		Dataset & Strategy & \#Param. & \#FLOPs & Top-1 \\
		\hline \hline
		\multirow{7}{*}{CUB-200} & VGG-16 & 135.07M & 30.93B & 72.30\% \\
		& FT \& prune & 7.91M & 9.34B & 66.90\% \\
		& Train from scratch & 7.91M & 9.34B & 44.27\% \\ 
		\cline{2-5}
		& ThiNet-Conv & 128.16M & 9.58B & 70.90\% \\
		& ThiNet-GAP & 7.91M & 9.34B & 69.43\% \\
		\cline{2-5}
		& ThiNet-Tiny & 1.12M & 2.01B & 65.45\% \\
		& AlexNet & 57.68M & 1.44B & 57.28\% \\
		\hline \hline
		\multirow{7}{*}{Indoor-67} & VGG-16 & 134.52M & 30.93B & 72.46\% \\
		& FT \& prune & 7.84M & 9.34B & 64.70\% \\
		& Train from scratch & 7.84M & 9.34B & 38.81\% \\
		\cline{2-5} 
		& ThiNet-Conv & 127.62M & 9.57B & 72.31\% \\
		& ThiNet-GAP & 7.84M & 9.34B & 70.22\% \\
		\cline{2-5}
		& ThiNet-Tiny & 1.08M & 2.01B & 62.84\% \\
		& AlexNet & 57.68M & 1.44B & 59.55\% \\
		\hline
	\end{tabular}
\end{table}

We first fine-tune the pre-trained VGG-16 model on the new dataset, which is a popular strategy adopted in numerous recognition tasks. As we can see, the fine-tuned model has the highest accuracy at the cost of huge model size and slow inference speed. Then, we use the proposed ThiNet approach to prune some unimportant filters (denoted by ``FT \& prune''), converting the cumbersome model into a much smaller one. With small-scale training examples, the accuracy cannot be recovered completely, \ie, the pruned model can be easily trapped into bad local minima. However, if we train a network from scratch with the same structure, its accuracy can be much lower. 

We suggest to fine-tune the ThiNet model, which is first pruned using the ImageNet data. As shown in Table~\ref{domain_adaption}, this strategy gets the best trade-off between model size and classification accuracy. It is worth noting that the ThiNet-Conv model can even obtain a similar accuracy as the original VGG-16, but is smaller and much faster.

We also report the performance of ThiNet-Tiny on these two datasets. Although ThiNet-Tiny has the same level of accuracy as AlexNet on ImageNet, it shows much stronger generalization ability. This tiny model can achieve $3\%\sim8\%$ higher classification accuracy than AlexNet when transferred into domain-specific tasks with 50$\times$ fewer parameters. And its model size is small enough to be deployed on resource constrained devices.

\section{Conclusion}

In this paper, we proposed a unified framework, namely ThiNet, for CNN model acceleration and compression. The proposed filter level pruning method shows significant improvements over existing methods.

In the future, we would like to prune the projection shortcuts of ResNet. An alternative method for better channel selection is also worthy to be studied. In addition, extensive exploration on more vision tasks (such as object detection or semantic segmentation) with the pruned networks is an interesting direction too. The pruned networks will greatly accelerate these vision tasks.

\section*{Acknowledgements}
This work was supported in part by the National Natural Science Foundation of China under Grant No. 61422203.

\clearpage

{\small
\bibliographystyle{ieee}
\bibliography{egbib}
}

\end{document}